\documentclass[runningheads]{llncs}
\usepackage[T1]{fontenc}
\usepackage{graphicx}
\usepackage{times}
\usepackage{helvet}
\usepackage{courier}
\usepackage{amsmath}
\usepackage{algorithm}
\usepackage{algorithmic}
\usepackage{csquotes} 
\usepackage{color}
\usepackage{paralist}
\usepackage{amssymb}
\usepackage{indentfirst}
\usepackage{subfigure}
\usepackage{float}
\usepackage{multirow}
\usepackage{cite}
\usepackage{mathrsfs}

\usepackage{mathrsfs} 
\usepackage{amsfonts}
\usepackage{todonotes}
\usepackage{pgfplots} 
\usepackage{epstopdf}
\usepackage{epsfig}
\pgfplotsset{compat=newest}
\usepackage{color}
\definecolor{forestgreen}{RGB}{0,139,69}
\usepackage{caption}

\usepackage[switch]{lineno}

\usepackage{textcomp,booktabs}
\usepackage{amssymb}
\usepackage{pifont}

\usepackage{makecell}
\usepackage{color}
\definecolor{forestgreen}{RGB}{0,139,69}
\usepackage{xcolor}
\definecolor{citecolor}{HTML}{0071bc}
\definecolor{SeaGreen4}{RGB}{0,205,102} 
\definecolor{SlateBlue}{RGB}{106,90,205} 
\definecolor{DarkRed}{RGB}{178,34,34} 
\usepackage[colorlinks, linkcolor=red,  anchorcolor=DarkRed, citecolor=SeaGreen4]{hyperref} 
\begin{document}
\title{ EMRRG: Efficient Fine-Tuning Pre-trained X-ray Mamba Networks for Radiology Report Generation   
\thanks{Corresponding author: Xiao Wang (\email{xiaowang@ahu.edu.cn})}  
}

\author{
Mingzheng Zhang \inst{1} \and
Jinfeng Gao \inst{1} \and 
Dan Xu \inst{1} \and 
Jiangrui Yu \inst{1} \and 
Yuhan Qiao \inst{1} \and \\ 
Lan Chen \inst{2} \and 
Jin Tang \inst{1} \and 
Xiao Wang \inst{1} 
}


\authorrunning{Mingzheng Zhang et al.}
%
\institute{
School of Computer Science and Technology, Anhui University, Hefei 230601, China
\and 
School of Electronic and Information Engineering, Anhui University, Hefei 230601, China
}

%




\maketitle              
\begin{abstract}
X-ray image-based medical report generation (MRG) is a pivotal area in artificial intelligence that can significantly reduce diagnostic burdens for clinicians and patient wait times. Existing MRG models predominantly rely on Large Language Models (LLMs) to improve report generation, with limited exploration of pre-trained vision foundation models or advanced fine-tuning techniques. Mainstream frameworks either avoid fine-tuning or utilize simplistic methods like LoRA, often neglecting the potential of enhancing cross-attention mechanisms. Additionally, while Transformer-based models dominate vision-language tasks, non-Transformer architectures, such as the Mamba network, remain underexplored for medical report generation, presenting a promising avenue for future research. In this paper, we propose EMRRG, a novel X-ray report generation framework that fine-tunes pre-trained Mamba networks using parameter-efficient methods. Specifically, X-ray images are divided into patches, tokenized, and processed by an SSM-based vision backbone for feature extraction, with Partial LoRA yielding optimal performance. An LLM with a hybrid decoder generates the medical report, enabling end-to-end training and achieving strong results on benchmark datasets. Extensive experiments on three widely used benchmark datasets fully validated the effectiveness of our proposed strategies for the X-ray MRG. 
The source code of this paper will be released on \url{https://github.com/Event-AHU/Medical_Image_Analysis}. 
\end{abstract}

\section{Introduction} 

\begin{figure*}
\centering
\includegraphics[width=1\linewidth]{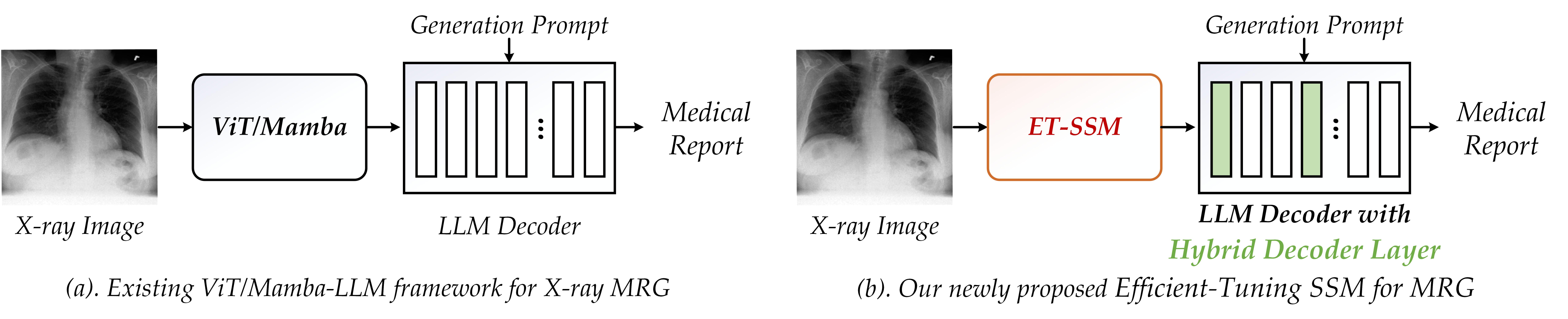}
\caption{Comparison between (a). Existing ViT/Mamba-LLM framework for X-ray medical report generation; 
(b). Our newly proposed Efficient-Tuning based SSM for MRG.} 
\label{fig:firstIMG}
\end{figure*}

X-ray based medical report generation (MRG)~\cite{rehman2025advancement} represents a pivotal application of artificial intelligence in healthcare, utilizing sophisticated AI models to generate high-fidelity diagnostic reports from radiographic images autonomously. This technology holds significant promise for mitigating clinical workload burdens, reducing diagnostic delays, and facilitating the operational integration of AI systems into routine clinical workflows. While AI performance has demonstrated remarkable progress, MRG systems continue to underperform relative to the diagnostic acumen of experienced clinicians, attributable to persistent technical and systemic limitations. The challenging issues include image interpretation, data annotation, heterogeneity issues, consistency and standardization of reports, diversity and variability of diseases, interpretability of algorithms, etc.

To address these challenges, some researchers resort to pre-trained foundation models, such as Llama~\cite{grattafiori2024llama}, Qwen~\cite{qwen}, to further augment the overall performance. Specifically, 
Chen et al. propose the R2Gen~\cite{chen2020generating}  framework based on the Transformer network, and many subsequent works~\cite{Wang2023R2GenGPT, Chen2021R2GenCMN, wang2024r2gencsr} are designed by following this framework. 
Wang et al. propose a novel X-ray medical report generation framework called R2GenCSR~\cite{wang2024r2gencsr} that adopts the Mamba architecture as its backbone and mines contextual samples to guide large language models for high-quality report generation. 
Wang et al. introduce the MambaXray-VL~\cite{wang2025cxpmrg}, which adopts the Mamba as the vision encoder and the large language model as the text decoder.

Despite these great progresses, current MRG systems are still limited by the following issues: 
\textit{1).} Most existing Medical Report Generation (MRG) models focus primarily on leveraging Large Language Models (LLMs)~\cite{Wang2023R2GenGPT} to enhance report generation capabilities. Yet, few works explore the use of pre-trained vision foundation models to improve overall performance, particularly through lightweight fine-tuning of these models.
\textit{2).} Mainstream frameworks that employ LLMs for report generation typically either avoid fine-tuning altogether or apply simple strategies such as LoRA~\cite{hu2022lora} for fine-tuning the large models. While these approaches can yield decent results, they primarily exploit the self-attention mechanism and overlook the potential of expanding cross-attention capabilities. 
\textit{3).} Transformer-based models have recently dominated various vision-language tasks, however, exploration of non-Transformer architectures remains notably insufficient. Investigating how to leverage cutting-edge non-Transformer models, e.g., the Mamba network~\cite{gu2023mamba}, to enhance the quality of medical report generation is a promising and underexplored direction.

In this paper, we propose a novel X-ray report generation framework by efficiently tuning the pre-trained Mamba networks, termed EMRRG. As shown in Fig.~\ref{fig:firstIMG}, the key insight of this paper lies in exploring novel parameter-efficient approaches to fine-tune pre-trained non-Transformer models and better adapt them to LLMs to enhance the quality of generated medical reports. Specifically, given an X-ray image, we first partition it into image patches and map them into tokens. These tokens are then fed into an efficiently tuned SSM (State Space Model)-based vision backbone for feature extraction. During this process, we investigate multiple fine-tuning paradigms and find that Partial LoRA achieves the best performance. Subsequently, we employ an LLM equipped with a hybrid decoder layer as the decoder network to generate the corresponding medical report. The entire framework supports end-to-end training and achieves strong experimental results across multiple mainstream benchmark datasets. The detailed network architectures can be found in Fig.~\ref{fig:framework}.

To sum up, the main contributions of this paper can be summarized as the following three aspects:

1). We propose a novel Mamba-based X-ray medical report generation framework, termed EMRRG, based on parameter efficient fine-tuning. 

2). We introduce a hybrid decoder layer augmented large language model (LLM) for accurate report generation. 

3). Extensive experiments on three benchmark datasets (i.e., IU X-ray~\cite{demner2016iuxray}, MIMIC~\cite{johnson2019mimicCXR}, CheXpert Plus Dataset~\cite{chambon2024chexpertPLUS}) fully validated the effectiveness of our proposed strategies for the radiology report generation.

\section{Related Work} 

\subsection{Mamba Network} 

State Space Model (SSM)~\cite{wang2024SSMSurvey} serves as the foundation for Mamba, functioning as a linear-complexity approach for modeling long-range dependencies. Evolving from the original SSM to the Structured State Space sequence model (S4)~\cite{gu2023mamba}, and subsequently to Mamba and Mamba2~\cite{dao2024transformers}, these architectures have continuously advanced in efficiency and theoretical grounding. Mamba~\cite{gu2023mamba} introduces data-dependent selective state updates and hidden state expansion, enabling efficient sequence modeling with linear time complexity. Its strong performance across diverse domains highlights its ability to maintain computational efficiency while achieving accuracy comparable to Transformer-based models. In computer vision, Mamba has been adapted into the Vision Mamba (ViM)~\cite{vim} architecture. ViM employs pure Mamba layers with bidirectional scanning, achieving efficient and scalable visual representation learning. This linear-complexity design makes ViM suitable for large-scale pretraining and multi-modal integration. However, as model scales grow, fine-tuning such architectures efficiently becomes increasingly challenging, motivating research into Parameter-Efficient Fine-Tuning (PEFT) strategies.

\subsection{Parameter Efficient Fine-Tuning} 

Parameter-Efficient Fine-Tuning (PEFT) aims to adapt large pre-trained models to downstream tasks while updating only a small subset of parameters~\cite{2025State}. Typical PEFT methods fall into three categories: partial-based, addition-based, and prompt-based approaches~\cite{galim2024parameter}. In contrast to the Transformer domain, PEFT research for Mamba remains relatively limited~\cite{2024MambaPEFT}. Recent efforts have begun exploring selective and structured tuning within Mamba architectures. SDLoRA~\cite{galim2024parameter} combines selective dimension tuning with Low-Rank Adaptation (LoRA), updating only specific channels and states within SSM modules while applying LoRA to linear projections. Selective Visual Prompting (SVP)~\cite{2024Selective}, designed for Vision Mamba (ViM)~\cite{vim}, introduces a lightweight dual-path prompting mechanism Cross-Prompting for shared information propagation and Inner-Prompting for layer-specific adaptation. Similarly, Partial LoRA~\cite{2024MambaPEFT} applies LoRA selectively to Mamba’s linear layers, improving performance with minimal parameter overhead. Despite these advances, PEFT for Mamba in high-resolution medical imaging remains underexplored, motivating our investigation into efficient fine-tuning frameworks for radiographic report generation.

\subsection{X-ray Report Generation}

X-ray report generation has evolved through several architectural paradigms, including CNN-based, RNN-based, and Transformer-based frameworks~\cite{wang2024r2gencsr}. Early works such as Li et al.~\cite{2020Convolutional} integrated CNNs for visual feature extraction with RNNs for sequential text generation, while Jing et al.~\cite{2017On} introduced an LSTM-based pipeline~\cite{Hochreiter1997LSTM} that predicts pathological findings before report composition. The introduction of Transformers led to major breakthroughs, Chen et al.~\cite{Chen2021R2GenCMN} utilized self-attention to generate detailed diagnostic descriptions, and Wang et al.~\cite{wang2024XrayHDMAE} enhanced vision-language pretraining by adapting Vision Transformers (ViT)~\cite{2020An} with masked autoencoding objectives on medical images. Recent developments have emphasized clinical relevance and interpretability. DCL~\cite{Li2023DCL} integrates dynamic graph reasoning to encode domain knowledge; RGRG~\cite{Tanida2023RGRG} combines lesion detection with region-guided text generation; and HERGen~\cite{wang2024hergen} models temporal dependencies among historical reports. The latest methods, such as R2GenGPT~\cite{Wang2023R2GenGPT} and R2GenCSR~\cite{wang2024r2gencsr}, incorporate large language models (LLMs) or Mamba-based visual backbones to enhance contextual reasoning and feature discriminability. In summary, the field has progressed from foundational CNN/RNN frameworks to Transformer-based and Mamba-based architectures that emphasize efficiency, temporal reasoning, and clinical alignment. Building on this evolution, our work integrates the Mamba backbone with PEFT mechanisms to achieve efficient and accurate X-ray report generation.

\section{Methodology} 
In this section, we will first give an overview of our proposed EMRRG model, then, we will dive into the details of the Partial LoRA and the hybrid decoder layers. Finally, we introduce the loss function used in this work.

\begin{figure*} 
\centering
\includegraphics[width=\textwidth]{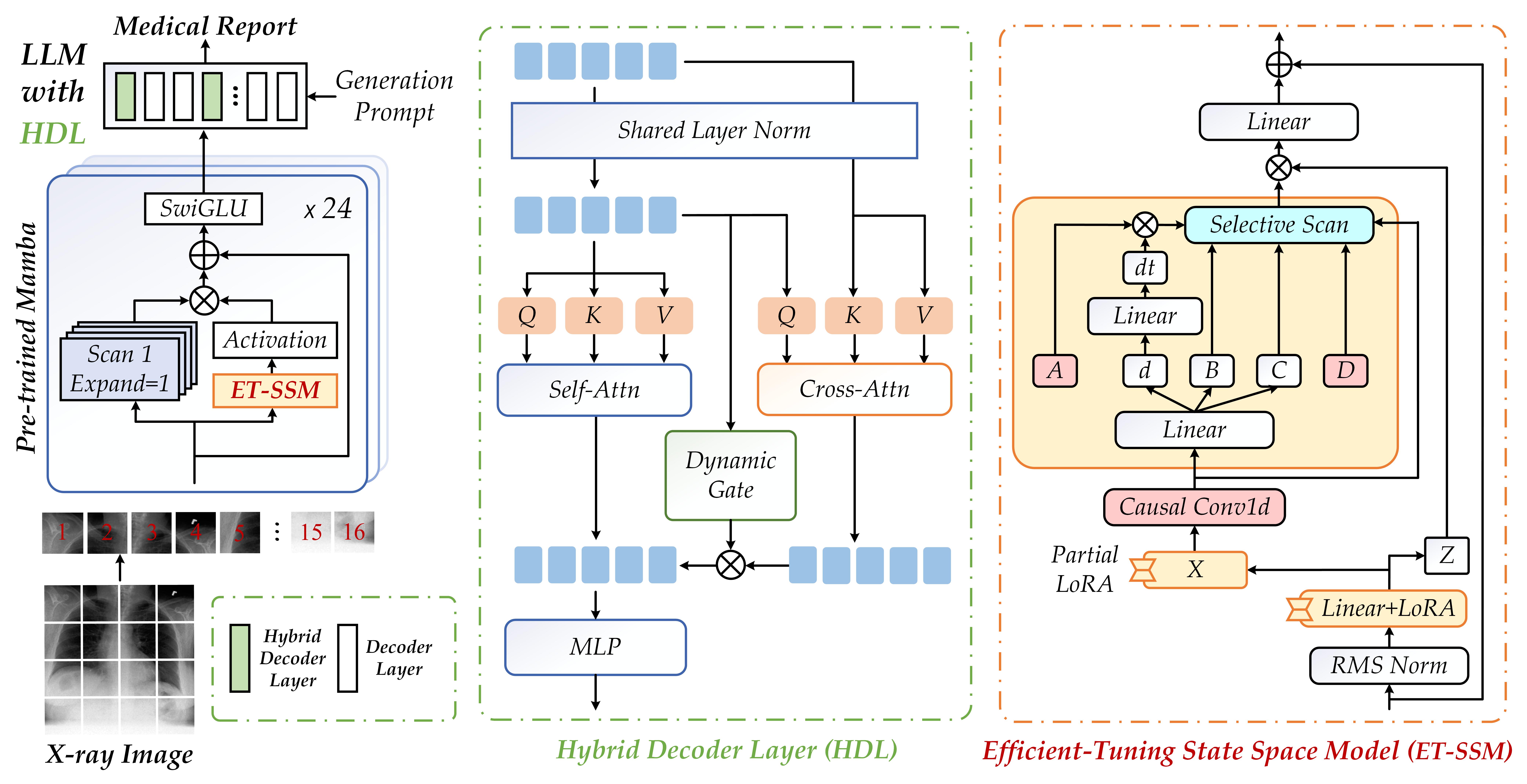} 
\caption{An overview of our proposed X-ray medical report generation framework by efficient tuning Mamba networks, termed EMRRG.} 
\label{fig:framework}
\end{figure*}

\subsection{Overview} 
As shown in Fig.~\ref{fig:framework}, we propose a novel framework for X-ray report generation. This framework comprises the efficient fine-tuning of pre-trained Mamba networks and the introduction of a hybrid decoder layer augmented large language models (LLMs) to facilitate accurate report generation. Compared to the Transformer, SSM models achieve higher computational efficiency owing to their sub-quadratic time complexity. Therefore, building upon the SSM architecture, we propose ET-SSM (Efficient-Tuning based SSM). This approach utilizes Partial LoRA to fine-tune the intermediate features 
\textit{X} of the Mamba model, while applying traditional LoRA to the input projection layer. This strategy enhances the quality of input representations at the feature extraction source and further improves training efficiency. More importantly, during the report generation stage, we replace part of the standard decoder layers with hybrid decoder layers incorporating a cross-attention mechanism, which demonstrates superior performance compared to the widely used self-attention mechanism.
Theoretically, the cross-attention mechanism allows text embeddings to dynamically extract relevant information from the uncompressed visual features, guided by the instruction to generate a diagnostic report. In the context of medical reporting, this enables the model to focus on key regions within the X-ray image (e.g., lesion sites), thereby producing more accurate and clinically relevant descriptions while avoiding interference from extraneous details.

\subsection{Input Encoding} 
In this paper, we denote the X-ray image as  \( I \in \mathbb{R}^{192 \times 192 \times 3} \), we initially partition it into non-overlapping image patches \( P_i \in \mathbb{R}^{16 \times 16 \times 3}, i = \{1, 2, ..., N\} \) and project them into visual tokens \( T_i \in \mathbb{R}^{1024}, i = \{1, 2, ..., N\} \)
using a convolutional layer with a kernel size of 16 × 16. Here, N is 144 when the input image resolution is set to 3 × 192 × 192. The visual tokens are subsequently fed into the Vim backbone network for feature extraction. This architecture achieves a significantly lower computational complexity of $\mathcal{O}(N)$ compared to the $\mathcal{O}(N^2)$ of the widely adopted Transformer. The core component of Vim is the Mamba block, a specific variation of the State Space Model (SSM), which enables this efficiency.

\subsection{Efficient-Tuning SSM} 
This work addresses the characteristic presence of numerous intermediate features with distinct properties (e.g., $X$, $Z$, $dt$, $B$, $C$) in Mamba networks by proposing an efficient fine-tuning strategy that integrates localized adaptation with global optimization. Although LoRA, as an architecture-agnostic parameter-efficient fine-tuning method, can be directly applied to Mamba, conventional implementations compress all intermediate features into a single low-rank subspace, thereby ignoring their inherent differences. As shown in the right part of Fig. \ref{fig:framework}, to tackle this limitation, we introduce the $\mathrm{LoRA}_{P}(X)$ \cite{2024MambaPEFT} method, which applies LoRA adaptations selectively to only a portion of the weights in linear layers based on the structure of the output features, achieving more refined parameter updates.

Furthermore, we employ conventional LoRA fine-tuning on the input projection layer to enhance the quality of input representations at the source of feature extraction. This approach not only strengthens the discriminative power of features subsequently processed by the selective scan mechanism but also facilitates synergistic modeling between the $X$ and $Z$ pathways. This comprehensive and hierarchical fine-tuning framework significantly improves the model's efficiency in the chest X-ray medical report generation task.

\subsection{LLM with Hybrid Decoder} 
The hybrid decoder layer constitutes the core architectural component of our LLM. As illustrated in the middle section of Fig. \ref{fig:framework}, this layer builds upon and extends the standard decoder layer. To ensure seamless integration of cross-attention within the decoder architecture, we conceptualize it as complementary to self-attention: whereas self-attention aggregates contextual information from preceding textual tokens, cross-attention simultaneously extracts relevant visual context from visual tokens.

The cross-attention module operates in parallel with self-attention, with its output being integrated back into the textual embeddings via a skip connection. A key innovation in this integration pathway is the introduction of a dynamic gating mechanism, which adaptively modulates the fused output prior to its incorporation into the main representation. This gating approach serves dual purposes: it effectively mitigates potential information interference while concurrently enhancing training stability. In the subsequent section, we provide a comprehensive elaboration of both the computational pipeline for cross-attention and the architectural design of the dynamic gate.

\noindent\textbf{Cross-attention.}
The hybrid decoder layer takes hidden states and visual tokens as input. Following prior works~\cite{ye2024mplug,shi2025slow}, these inputs are first normalized through a shared layer normalization. The normalized outputs then undergo two parallel attention mechanisms: while all hidden states participate in self-attention, only text-derived queries from the self-attention layer are utilized for cross-attention with visual tokens. For conciseness, attention head dimensions are omitted in our description. We denote the textual query by \(\mathbf{Q}_t\in\mathbb{R}^{n\times d}\), where \(n\) is the number of text tokens and \(d\) is the hidden dimension.
Let the (normalized) visual features be \(\mathbf{V}_s\in\mathbb{R}^{m\times d}\), with \(m\) visual tokens.
The cross-attention output \(\mathbf{X}'\in\mathbb{R}^{n\times d}\) is computed by a multi-head cross-attention (MHCA) module as
\begin{equation}
\mathbf{X}' \;=\; \operatorname{MHCA}\!\big(\mathbf{Q}_t,\; W_k\mathbf{V}_s,\; W_v\mathbf{V}_s\big),
\end{equation}
where \(\operatorname{MHCA}(q,k,v)\) denotes multi-head cross-attention with query \(q\), key \(k\), and value \(v\).
\(W_k\) and \(W_v\) are learnable linear projections that align visual features to the key and value spaces, respectively.


\noindent\textbf{Dynamic gate with warm-up.}
To mitigate potential noise introduced by cross-attention outputs $X'$ into pretrained LLMs, we employ a dynamic gating mechanism with warm-up initialization. While existing approaches typically use learnable scalar gates~\cite{grattafiori2024llama}, these static assignments fail to account for varying visual context requirements across different tokens, as demonstrated by divergent attention patterns~\cite{zhu2024focusllava}. Our method generates token-wise gating values $g_t$ through a linear layer followed by tanh activation, enabling adaptive fusion of visual features into textual embeddings. Following~\cite{shi2025slow}, we initialize a warm-up scalar to zero to stabilize early training phases, allowing the model to learn optimal integration weights throughout the multimodal alignment process gradually.
The output of the cross-attention module is then integrated into the textual embedding ${X}_t$ via a gated residual update:
\begin{equation}
\mathbf{X}_t \leftarrow \mathbf{X}_t + \big(\mathbf{X}' \odot g_d\big)\cdot g_s,
\end{equation}
where \(\odot\) denotes the element-wise (Hadamard) product, and \(g_d\) and \(g_s\) are gating factors that modulate the contribution of the cross-attention output. 
This gated residual connection updates the text-specific component of the hidden states, while the self-attention branch provides complementary contextual refinement across tokens. 


\subsection{Loss Function} 
In this paper, we adopt the negative log-likelihood as the loss function, i.e.,
\begin{equation}
\label{eq:nll}
\mathcal{L}_{NLL} = -\sum_{i=1}^{T} \log p_\theta(y_i|Prompt, [y_1,...,y_{i-1}]),
\end{equation}
where $\theta$  represents the trainable parameters, T is the number of words predicted by the Large Language Model, and Prompt is the instruction used in our experiments, i.e., "Generate a comprehensive and detailed diagnosis report for this chest X-ray image."

\section{Experiments} 

\subsection{Datasets and Evaluation Metric}  
In this article, three public benchmark datasets are adopted for experimental validation and analysis, including IU-Xray~\cite{demner2016iuxray}, MIMIC-CXR~\cite{johnson2019mimicCXR}, and CheXpert Plus~\cite{chambon2024chexpertPLUS} datasets. A brief introduction to these datasets is given below. 

\noindent $\bullet$ \textbf{IU X-ray Dataset} \cite{demner2016iuxray}
published in 2016, is one of the most frequently used publicly available medical image datasets for the generation of medical reports. It contains 7,470 images and 3,955 radiology reports; each report is associated with either a frontal view image or both frontal and lateral view images. Each report is divided into four sections: Indication, Comparison, Findings, and Impression. For a fair comparison, we used the same dataset split protocol as R2GenGPT \cite{Wang2023R2GenGPT}, dividing the dataset into training, testing, and validation sets with a ratio of 7:1:2.

\noindent $\bullet$ \textbf{MIMIC-CXR Dataset} \cite{johnson2019mimicCXR}
is one of the largest publicly available chest X-ray datasets, containing free-text radiology reports. These records from 2011-2016 include 377,110 radiographic images and 227,835 radiology reports collected from 65,379 patients at the Beth Israel Deaconess Medical Center Emergency Department in Boston, Massachusetts. For fair comparison, we used the same dataset split protocol as R2GenGPT, with 270,790 samples for training the model, and 2,130 and 3,858 samples for validation and testing sets, respectively.

\noindent $\bullet$ \textbf{CheXpert Plus Dataset} \cite{chambon2024chexpertPLUS} 
is a new radiology dataset designed to enhance the scale, performance, robustness, and fairness of deep learning models in the field of radiology. This dataset includes 223,228 chest X-rays (in DICOM and PNG formats), 187,711 corresponding radiology reports (de-identified and parsed into 11 sections), de-identified demographic data from 64,725 patients, 14 chest pathology labels, and RadGraph \cite{jain2021RadGraph} annotations. For a fair comparison, we followed the dataset split protocol used in R2GenCSR \cite{wang2024r2gencsr}, which adopted Findings as the ground truth and split the training/validation/testing subsets based on the ratio 7:1:2. The training subset contains 40,463 samples, the validation subset contains 5,780 samples, and the testing subset contains 11,562 samples.

For evaluation metrics, we adopt natural language metrics and clinical metrics to evaluate our generated X-ray reports. For the natural language metrics, we choose BLEU\cite{papineni_bleu_2002}, ROUGE-L \cite{lin_rouge_2004}, METEOR \cite{banerjee_meteor_2005}, and CIDEr \cite{vedantam_cider_2015}. For the clinical metrics, i.e., \textit{Precision}, \textit{Recall}, and \textit{F1-measure}, the formula can be formally defined as: 
\begin{align}
\label{PRFmetric} 
Precision & = \frac{TP}{TP+FP}, \\ 
Recall & = \frac{TP}{TP+FN}, \\ 
F1 & = \frac{2\times P \times R}{P+R}
\end{align}
where TP (True Positive) refers to instances that are correctly identified as positive, FP (False Positive) denotes cases incorrectly labeled as positive when they are actually negative (also known as a Type I error), and FN (False Negative) represents instances incorrectly classified as negative despite being positive (referred to as a Type II error). The F1 Score serves as a harmonic mean of Precision and Recall, providing a balanced measure that accounts for both the accuracy of positive predictions and the ability to capture all relevant positive cases, without favoring one metric over the other.

\subsection{Implementation Details}  
We tested the model's performance on three widely used public benchmark datasets. On the IU-Xray \cite{demner2016iuxray} dataset, we set the maximum training epochs to 30 and the batch size to 20. The visual encoder used was Vim \cite{vim}, loaded with pre-trained weights from~\cite{wang2025cxpmrg}, while the large language model was Qwen-1.5-1.8B \cite{qwen}, with max\_length set to 60 and a validation frequency of 1, meaning we validated after each training epoch. On the MIMIC-CXR  \cite{johnson2019mimicCXR} and CheXpert Plus  \cite{chambon2024chexpertPLUS} datasets, we set the maximum training epochs to 6 and the batch size to 18. The visual encoder remained unchanged, while the large language model used was Llama2-7B \cite{touvron2023llama2}, with max\_length set to 100 and a validation frequency of 0.5, meaning we validated at both the end of each training cycle and after the training was complete. When implementing the $\mathrm{LoRA}_{P}(X)$  methodology for fine-tuning, specifically during the application of LoRA to partial weights of the in\_proj layer, the rank parameter $r$ was configured to 32. For the language model, four hybrid decoder layers are distributed within the LLM's decoder layers at indices [0, 8, 16, 24]. More details can be found in our source code.

\begin{table*}[h!]
\caption{ Comparison with SOTA on the \textbf{NLG Metrics}. The symbol $\dagger$ indicates that we follow the R2Gen annotation using \textit{Findings} and evaluate with our method, as their report modifies the ground truth to an \textit{Impression} concatenated with \textit{Findings}. The best result is highlighted in bold, and the second-best result is underlined. }
\label{tab:NLG_Metrics_results}
\resizebox{\linewidth}{!}{
\begin{tabular}{c|l|l|ccccccc}
\hline \toprule [0.5 pt] 
\textbf{Dataset} & \textbf{Method} & \textbf{Publication} & \textbf{BLEU-1} & \textbf{BLEU-2} & \textbf{BLEU-3} & \textbf{BLEU-4} & \textbf{ROUGE-L} & \textbf{METEOR} & \textbf{CIDEr} \\ \hline
\multirow{10}{*}{\textbf{IU X-Ray}} 
 & R2Gen~ \cite{chen2020generating} & EMNLP 2020 & 0.470 & 0.304 & 0.219 & 0.165 & 0.371 & 0.187 & - \\
 & R2GenCMN~ \cite{Chen2021R2GenCMN} & ACL-IJCNLP 2021 & 0.475 & 0.309 & 0.222 & 0.170 & 0.375 & 0.191 & - \\
 & METransformer~ \cite{Wang2023METransformer} & CVPR 2023 & 0.483 & 0.322 & 0.228 & 0.172 & 0.380 & 0.192 & 0.435 \\
 & DCL~ \cite{Li2023DCL} & CVPR 2023 & - & - & - & 0.163 & \underline{0.383} & 0.193 & \textbf{0.586} \\
 & R2GenGPT\textsuperscript{$\dagger$}~ \cite{Wang2023R2GenGPT} & Meta Radiology 2023 & 0.465 & 0.299 & 0.214 & 0.161 & 0.376 & 0.219 & 0.542 \\
 & PromptMRG~ \cite{Jin2024PromptMRG} & AAAI 2024 & 0.401 & - & - & 0.098 & 0.160 & \textbf{0.281} & - \\  
 & Med-LMM~ \cite{Liu2024Med_LMM} & ACM MM 2024 & - & - & - & 0.168 & 0.381 & 0.209 & 0.427 \\ 
 & SILC~ \cite{liu2024multi} & IEEE TMI 2024 & 0.472 & \underline{0.321} & \underline{0.234} & \underline{0.175} & 0.379 & 0.192 & 0.368 \\ 
& KIA~ \cite{yin2025kia} & COLING 2025 & \textbf{0.501} & \textbf{0.325} & \textbf{0.240} & \textbf{0.183} & 0.375 & 0.207 & \underline{0.559} \\ 
 \cline{2-10} 
 & EMRRG & Ours & \underline{0.487} &  \textbf{0.325} & 0.222 & 0.167 & \textbf{0.385} & \underline{0.226} &  0.476 \\
 \hline \toprule [0.5 pt] 
\multirow{10}{*}{\textbf{MIMIC-CXR}} 
 & R2Gen~ \cite{chen2020generating} & EMNLP 2020 & 0.353 & 0.218 & 0.145 & 0.103 & 0.277 & 0.142 & - \\
 & R2GenCMN~ \cite{Chen2021R2GenCMN} & ACL-IJCNLP 2021 & 0.353 & 0.218 & 0.148 & 0.106 & 0.278 & 0.142 & - \\
 & METransformer~ \cite{Wang2023METransformer} & CVPR 2023 & 0.386 & 0.250 & 0.169 & 0.124 & \underline{0.291} & 0.152 & \textbf{0.362} \\
 & DCL~ \cite{Li2023DCL} & CVPR 2023 & - & - & - & 0.109 & 0.284 & 0.150 & \underline{0.281} \\
 & R2GenGPT\textsuperscript{$\dagger$}~ \cite{Wang2023R2GenGPT} & Meta Radiology 2023 & \textbf{0.408} & \textbf{0.256} & 0.174 & 0.125 & 0.285 & \textbf{0.167} & 0.244 \\
 & PromptMRG~ \cite{Jin2024PromptMRG} & AAAI 2024 & 0.398 & - & - & 0.112 & 0.268 & 0.157 & - \\ 
 & Med-LMM~ \cite{Liu2024Med_LMM} & ACM MM 2024 & - & - & - & \underline{0.128} & 0.289 & 0.161 & 0.265 \\ 
 & AdaMatch-Cyclic~ \cite{chen2024AdaMatch_Cyclic} & ACL 2024 & 0.379 & 0.235 & 0.154 & 0.106 & 0.286 & 0.163 & - \\
& GIT-CXR~ \cite{sirbu2025git} & arXiv 2025 & 0.403 & \underline{0.254} & \textbf{0.215} & \textbf{0.136} & \textbf{0.311} & 0.161 & - \\
 \cline{2-10} 
 & EMRRG & Ours & \underline{0.407} & \textbf{0.256} & \underline{0.175} & 0.125 & 0.288 & \underline{0.164} & 0.239 \\ 
\hline  \toprule [0.5 pt] 
\multirow{10}{*}{\textbf{CheXpert Plus}} 
 & R2Gen~ \cite{chen2020generating} & EMNLP 2020 & 0.301 & 0.179 & 0.118 & 0.081 & 0.246 & 0.113 & 0.077 \\
 & R2GenCMN~ \cite{Chen2021R2GenCMN} & ACL-IJCNLP 2021 & 0.321 & 0.195 & 0.128 & 0.087 & 0.256 & 0.127 & 0.102 \\
 & XProNet~ \cite{wang2022cross} & ECCV 2022 & 0.364 & 0.225 & 0.148 & 0.100 & 0.265 & \underline{0.146} & 0.121 \\
 & ORGan~ \cite{hou2023ORGan} & ACL 2023 & 0.320 & 0.196 & 0.128 & 0.086 & 0.261 & 0.135 & 0.107 \\
 & R2GenGPT~ \cite{Wang2023R2GenGPT} & Meta Radiology 2023 & 0.361 & 0.224 & \underline{0.149} & \underline{0.101} & \underline{0.266} & 0.145 & \underline{0.123} \\
 & ASGMD~ \cite{XUE2024ASGMD} &ESWA 2024 & 0.267 & 0.149 & 0.094 & 0.063 & 0.220 & 0.094 & 0.044 \\
 & Token-Mixer~ \cite{yang2024token} &IEEE TMI 2024 & \textbf{0.378} & \underline{0.231} & \textbf{0.153} & 0.091 & 0.262 & 0.135 & 0.098 \\
 & PromptMRG~ \cite{Jin2024PromptMRG} & AAAI 2024 & 0.326 & 0.174 & - & 0.095 & 0.222 & 0.121 & 0.044 \\ 
 & R2GenCSR~ \cite{wang2024r2gencsr} & arXiv 2024 & 0.364 & 0.225 & 0.148 & 0.100 & 0.265 & \underline{0.146} & 0.121 \\
 \cline{2-10} 
 & EMRRG & Ours & \underline{0.375} & \textbf{0.232} & \textbf{0.153} & \textbf{0.104} & \textbf{0.273} & \textbf{0.152} & \textbf{0.167} \\ 
\hline  \toprule [0.5 pt] 
\end{tabular}
}
\end{table*}

\subsection{Comparison on Public Benchmarks} 
\noindent $\bullet$ \textbf{Results on IU X-ray Dataset.}~ 
As shown in Table \ref{tab:NLG_Metrics_results}, our EMRRG exhibits excellent performance on the IU X-ray dataset. To elaborate further, the EMRRG model is at the SOTA level on BLEU-2 (B2) and ROUGEL (R) metrics with scores of 0.325 and 0.385, respectively. Furthermore, we attained the second-best results on the BLEU-1 (B1) and METEOR (M) metrics, with corresponding scores of 0.487 and 0.226, respectively. This result demonstrates that our method achieves performance comparable to other report generation approaches. However, on some other metrics, such as BLEU-3 (B3),  BLEU-4 (B4), and CIDEr (C), our method does not achieve comparable performance. This reflects the need to improve the generalization of our method on other datasets.

\noindent $\bullet$ \textbf{Results on MIMIC-CXR Dataset.}~ As shown in Table~\ref{tab:NLG_Metrics_results}, our method also demonstrates outstanding performance on the MIMIC-CXR dataset, and achieves the most comparable level in several common indicators (e.g., BLEU-1, BLEU-2, BLEU-3, and METEOR). Specifically, our method outperforms R2GenGPT on the ROUGE-L metric and also achieves favorable performance on the BLEU-4 metric. In the CIDEr metric, our model achieved a score of 0.239, indicating that EMRRG still has room for improvement. Furthermore, as shown in Table \ref{tab:CE_Metrics_Mimic_CXR_results}, our method also demonstrates comparable performance on the CE metric; our scores
on Precision (P), Recall (R), and F1-score (F1) are 0.421,0.372, and 0.395, respectively.

\noindent $\bullet$ \textbf{Results on CheXpert Plus Dataset.}~ As shown in Table~\ref{tab:NLG_Metrics_results} and Table~\ref{tab:CE_Metrics_CheXpert_Plus_results}, our model EMRRG achieves  state-of-the-art performance across nearly all evaluation metrics.These include NLG evaluation metrics and CE evaluation metrics. In detail, for the NLG metrics, our scores on BLEU-4, ROUGE-L, METEOR, and CIDEr are 0.104, 0.273, 0.152, and 0.167, respectively. For the CE metrics, our scores on Precision (P), Recall (R), and F1-score (F1) are 0.341, 0.273, and 0.272, respectively. These experimental results fully demonstrate the superior performance of our model.

\begin{table*}[t]
\centering
\small
\begin{minipage}[t]{0.48\linewidth}
\centering
\caption{ Comparison of the \textbf{CE Metrics} on MIMIC-CXR dataset. } 
\label{tab:CE_Metrics_Mimic_CXR_results}
\resizebox{\linewidth}{!}{
\begin{tabular}{l|c|ccc}
\hline \toprule [0.5 pt] 
\multirow{2}{*}{\textbf{Method}} & \multirow{2}{*}{\textbf{Publication}} & \multicolumn{3}{c}{\textbf{MIMIC-CXR}}  \\ \cline{3-5} 
 & & \textbf{Precision} &  \textbf{Recall} &  \textbf{F1}  \\ \hline
 R2Gen~ \cite{chen2020generating} & EMNLP 2020  & 0.333 & 0.273 & 0.276   \\
 METransformer~ \cite{Wang2023METransformer} & CVPR 2023  & 0.364 & 0.309 & 0.311 \\
 KiUT~ \cite{huang2023kiut} & CVPR 2023  & 0.371 & 0.318 & 0.321 \\
 DCL~ \cite{Li2023DCL} & CVPR 2023  & \underline{0.471} & 0.352 & 0.373 \\
 CoFE~ \cite{li2024contrastive} & ECCV 2024 & \textbf{0.489} & 0.370 & \textbf{0.405}  \\
 HERGen~ \cite{wang2024hergen} & ECCV 2024  & 0.415 & 0.301 & 0.317 \\
 SILC~ \cite{liu2024multi} & IEEE TMI 2024  & 0.457 & 0.337 & 0.330 \\
 OaD~ \cite{li2024organ} & IEEE TMI 2024 & 0.364 & \underline{0.382} & 0.372  \\
 GIT-CXR~ \cite{sirbu2025git} & arXiv 2025 &
 0.349 & \textbf{0.403} & 0.336 \\
\hline 
 EMRRG & Ours & 0.421 & 0.372 & \underline{0.395}  \\
\hline \toprule [0.5 pt] 
\end{tabular}
}
\end{minipage}
\hfill
\begin{minipage}[t]{0.48\linewidth}
\centering
\caption{ Comparison of the \textbf{CE Metrics} on CheXpert Plus dataset. } 
\label{tab:CE_Metrics_CheXpert_Plus_results}
\resizebox{\linewidth}{!}{
\begin{tabular}{l|c|ccc}
\hline \toprule [0.5 pt] 
\multirow{2}{*}{\textbf{Method}} & \multirow{2}{*}{\textbf{Publication}} & \multicolumn{3}{c}{\textbf{CheXpert Plus}}  \\ \cline{3-5} 
 & & \textbf{Precision} &  \textbf{Recall} &  \textbf{F1}  \\ \hline
 R2Gen~ \cite{chen2020generating} & EMNLP 2020  & 0.318 & 0.200 & 0.181   \\
 R2GenCMN~ \cite{Chen2021R2GenCMN} & ACL 2021 & 0.329 & 0.241 & 0.231 \\
 XProNet~ \cite{wang2022cross} & ECCV 2022  & 0.314 & 0.247 & 0.259 \\
 R2GenGPT~ \cite{Wang2023R2GenGPT} & Meta-Rad. 2023  & 0.315 & 0.244 & 0.260 \\
 Zhu et al.~ \cite{zhu2023utilizing} & MICCAI 2023 & 0.217 & \textbf{0.308} &  0.205 \\
 PromptMRG~ \cite{Jin2024PromptMRG} & AAAI 2024  & 0.258 & 0.265 & \underline{0.281} \\
 Token-Mixer~ \cite{yang2024token} & IEEE TMI 2024  & 0.309 & 0.270 & \textbf{0.288} \\
\hline 
 EMRRG & Ours &  \textbf{0.341} & \underline{0.273} & 0.272  \\
\hline \toprule [0.5 pt] 
\end{tabular}
}
\end{minipage}
\end{table*}

\begin{table*}[h!]
\centering
\small
\caption{Runtime Efficiency on the CheXpert Plus Dataset using \textbf{Mainstream Medical Report Generation Algorithms}. Min is short for minutes. The Param listed in this table denotes the parameters that need to be tuned in the training phase.}
\resizebox{\linewidth}{!}{
\label{tab:resource_comparison}
\begin{tabular}{@{}c|l|l|l|l|c|c@{}}
\toprule
\textbf{Index} & \textbf{Algorithm} & \textbf{Publish} & \textbf{Encoder} & \textbf{Decoder} & \textbf{Time (min)} & \textbf{Param (M)} \\ 
\hline
\#01 & TIMER \cite{wu2023token} & CHIL23 & Transformer & Transformer & 26.5 & 79.28 \\ \hline
\#02 & CvT2DistilGPT2 \cite{nicolson2023improving} & AIM23 & Transformer & GPT2 & 13.93 & 128.00 \\ \hline
\#03 & ORGan \cite{hou2023ORGan}  & ACL23 & CNN & Transformer & 46.66 & 67.50 \\ \hline
\#04 & Zhu et al. \cite{zhu2023utilizing} & MICCAI23 & Transformer & Transformer & 10.03 & 85.95 \\ \hline
\#05 & CAMANet \cite{wang2024camanet} & IEEE JBH23 & Swin-Former & Transformer & 23.08 & 43.22 \\ \hline
\#06 & ASGMD \cite{XUE2024ASGMD} & ESWA24 & ResNet-101 & Transformer & 87.37 & 277.41 \\ \hline
\#07 & Token-Mixer \cite{yang2024token} & IEEE TMI23 & ResNet-50 & Transformer & 17.54 & 104.34 \\ \hline
\#08 & PromptMRG \cite{Jin2024PromptMRG} & AAAI24 & ResNet-101 & Bert & 108.45 & 219.92 \\ \hline
\#09 & R2GenGPT \cite{Wang2023R2GenGPT} & Meta-Rad.23 & Swin-Transformer & Llama2 & 77.80 & 90.90 \\ \hline
\#10 & R2GenCSR \cite{wang2024r2gencsr} & arXiv24 & VMamba & Llama2 & 31.20 & 91.70 \\ \hline
\#11 & Wang et al. \cite{wang2025cxpmrg} & arXiv24 & ViT & Llama2 & 10.82 & 358.80 \\ \hline
\#12 & MambaXray-VL \cite{wang2025cxpmrg} & CVPR2025 & MambaXray-VL & Llama2 & 50.66 & 57.31 \\ \hline
\#13 & EMRRG & Ours & MambaXray-VL & Llama2 & 26.84 & 1.32 \\
\bottomrule
\end{tabular}
} 
\end{table*}

\newcommand{\greencheck}{{\color{green}$\checkmark$}}
\newcommand{\redcross}{{\color{red}$\times$}}

\begin{table*}[h!]
\centering
\small 
\caption{Component analysis of the key method in our framework on the IU X-ray dataset. It is stated as follows: full fine-tuning (FT), Llama2 (L2), hybrid decoder layer (HDL). The best result is highlighted in bold, and the second-best result is underlined.} 
\label{tab:ablation_results}
\begin{tabular}{c|cc|ccc|ccccccc}
\hline
\multirow{2}{*}{Index} & \multicolumn{2}{c|}{SSM} & \multicolumn{3}{c|}{LLM} & \multicolumn{7}{c}{IU X-ray} \\
\cline{2-13}
 & FT & $\mathrm{LoRA}_{P}(X)$ & L2 & L2+LoRA & L2+HDL & B1 & B2 & B3 & B4 & R-L & M & C \\
\hline
\#01 & \greencheck & \redcross & \greencheck & \redcross & \redcross & 0.480 & 0.322 & \underline{0.226} & \underline{0.175} & 0.383 & 0.215 & 0.478 \\
\hline
\#02 & \greencheck & \redcross & \redcross & \greencheck & \redcross & 0.473 & 0.311 & 0.216 & 0.171 & 0.384 & 0.210 & \underline{0.483} \\
\#03 & \greencheck & \redcross & \redcross & \redcross & \greencheck & \textbf{0.489} & \underline{0.324} & \textbf{0.231} & \textbf{0.182} & \textbf{0.391} & \underline{0.219} & \textbf{0.490} \\
\#04 & \redcross & \greencheck & \greencheck & \redcross & \redcross & 0.475 & 0.309 & 0.211 & 0.161 & 0.372 & 0.208 & 0.464 \\
\#05 & \redcross & \greencheck & \redcross & \greencheck & \redcross & 0.471 & 0.313 & 0.216 & 0.158 & 0.374 & 0.210 & 0.461 \\
\#06 & \redcross & \greencheck & \redcross & \redcross & \greencheck & \underline{0.487} & \textbf{0.325} & 0.222 & 0.167 & \underline{0.385} & \textbf{0.226} & 0.476 \\
\hline
\end{tabular}
\end{table*}

\vspace{0.5cm}
\noindent

\normalsize

\begin{table}[h]
\centering
\small 
\caption{NLG Metrics of EMRRG Framework Components Fine-tuned with LoRA and Partial LoRA on IU X-ray Dataset. The best result is highlighted in bold, and the second-best result is underlined.}
\begin{tabular}{l|ccccccc}
\toprule
Setting   & BLEU-1 & BLEU-2 & BLEU-3 & BLEU-4 & ROUGE-L & METEOR & CIDEr \\
\midrule
LoRA(Llama2) & \underline{0.479} & \textbf{0.320} & \textbf{0.233} & \textbf{0.177} & 0.386 & \underline{0.212} & \underline{0.489} \\
LoRA(embedding) & 0.463 & 0.302 & 0.218 & 0.167 & 0.379 & 0.207 & \textbf{0.530} \\
LoRA(x\_proj) & 0.457 & 0.301 & \underline{0.226} & 0.162 & 0.383 & 0.205 & 0.473 \\
LoRA(dt\_proj) & 0.458 & 0.300 & 0.212 & 0.161 & 0.378 & 0.203 & 0.485 \\
LoRA(in\_proj) & 0.444 & 0.288 & 0.205 & 0.152 & 0.368 & 0.194 & 0.431 \\
LoRA(out\_proj) & 0.456 & 0.297 & 0.211 & 0.156 & 0.374 & 0.199 & 0.451 \\
LoRA$_{p}$(Z) & 0.466 & 0.302 & 0.213 & 0.163 & 0.381 & 0.204 & 0.467 \\
LoRA$_{p}$(dt) & 0.458 & 0.299 & 0.221 & 0.157 & \underline{0.387} & 0.197 & 0.484 \\
LoRA$_{p}$(B) & 0.462 & 0.307 & 0.208 & 0.165 & 0.384 & 0.201 & 0.472 \\
LoRA$_{p}$(C) & 0.473 & 0.303 & 0.217 & 0.154 & 0.379 & 0.209 & 0.466 \\
LoRA$_{p}$(X) & \textbf{0.485} & \underline{0.311} & 0.223 & \underline{0.169} & \textbf{0.388} & \textbf{0.216} & 0.474 \\
\bottomrule
\end{tabular}
\label{tab:lora_results}
\end{table}

\subsection{Ablation Study} 

\noindent $\bullet$ \textbf{Component Analysis.~} 
In Table \ref{tab:ablation_results}, for the SSM,  $\mathrm{LoRA}_{P}(X)$ indicates fine-tuning the intermediate feature \textit{X} of the SSM using the Partial LoRA method. 
For the LLM, L2+LoRA represents fine-tuning Llama2 with the conventional LoRA method; L2+HDL represents replacing part of the original decoder layers in Llama2 with hybrid decoder layers.
B1--B4 are short for BLEU-1 to BLEU-4, respectively, and R-L, M, and C are short for ROUGE-L, METEOR, and CIDEr, respectively.

From the results presented in Table \ref{tab:ablation_results}, under the condition of keeping the LLM operations unchanged, the $\mathrm{LoRA}_{P}(X)$ method achieves performance close to full fine-tuning (FT) across multiple metrics. Notably, for certain metrics, the $\mathrm{LoRA}_{P}(X)$ method even outperforms FT. As demonstrated by comparing rows \#02 and \#05, the $\mathrm{LoRA}_{P}(X)$ method performs comparably to FT on BLEU-3 and METEOR metrics, while surpassing FT on the BLEU-2 metric. When maintaining consistent SSM operations, the Llama2+HDL approach demonstrates superior performance over both standard Llama2 and Llama2+LoRA across nearly all evaluation metrics.

Overall, while the combination of the FT method with Llama2+HDL achieves optimal results on multiple metrics, and the $\mathrm{LoRA}_{P}(X)$ method with Llama2+HDL only attains optimal performance on BLEU-2 and METEOR metrics, our proposed EMRRG framework still maintains performance close to the full parameter fine-tuning approach. Importantly, it should be emphasized that our EMRRG framework requires training only 2.3\% of the parameters compared to full fine-tuning methods, resulting in significantly higher training efficiency than full parameter fine-tuning approaches.

\noindent $\bullet$ \textbf{Analysis of Different Tuning Settings.~} 
As shown in Table~\ref{tab:lora_results}, the results of fine-tuning different components using the LoRA method indicate that fine-tuning the parameters of the decoder Llama2 achieves superior performance, obtaining optimal scores on the BLEU-2, BLEU-3, and BLEU-4 metrics. This inspires further in-depth research on the text decoder, specifically by replacing part of the original decoder layers with hybrid decoder layers. Moreover, from the fine-tuning results of intermediate feature networks with various attributes in Mamba (such as X, Z, dt, B, C, etc., see Figure~\ref{fig:framework}) using the Partial LoRA method, \(\mathrm{LoRA}_{P}(X)\) attains the best performance across multiple metrics, thereby laying the experimental foundation for our proposed ET-SSM framework.

\begin{figure}[t]
    \centering
    \includegraphics[width=\linewidth]{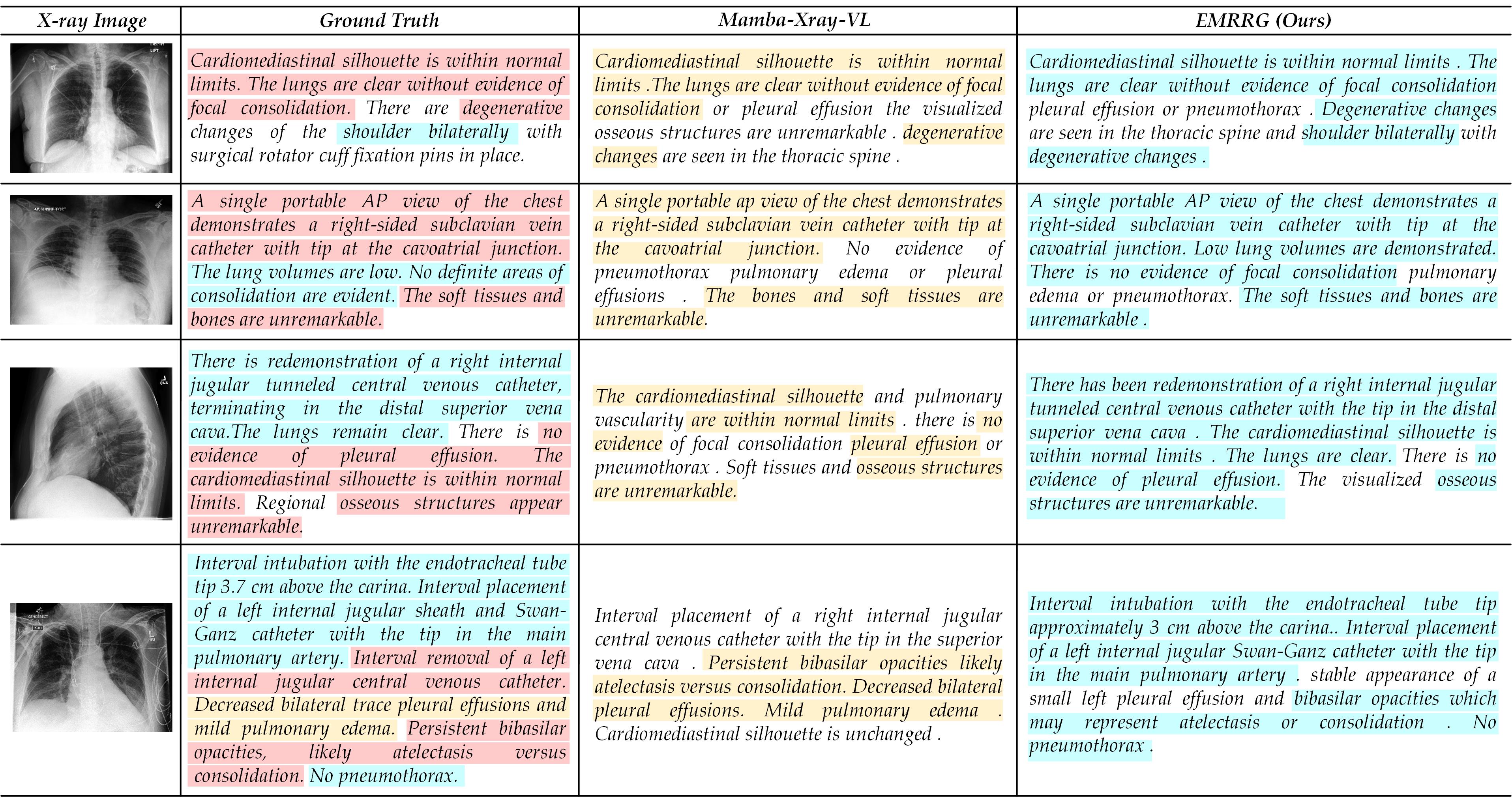}
    \caption{
        Comparison of generated reports on the CheXpert dataset. 
        Yellow highlights indicate sentences in the \textbf{MambaXray-VL} report matching the ground truth,   
        cyan highlights denote matching sentences in our \textbf{EMRRG} report, 
        and pink highlights represent matches shared by both models.
    }
    \label{fig:visualization}
\end{figure}

\noindent $\bullet$ \textbf{Efficiency Analysis}
From the perspective of running efficiency, we test these models on a server with A800 GPUs (80GB). Note that we set the batch size as large as possible to make full use of the GPU memory. As a result, we can find that MSAT \cite{wang2024camanet} and XProNet \cite{wang2022cross} are the first two algorithms that only need 5.72 and 6.3 minutes for the testing subset. R2Gen \cite{Chen2021R2GenCMN}, PromptMRG \cite{Jin2024PromptMRG}, and VLCI \cite{Liu_2023_10} are relatively slow and need more than 100 minutes on the testing subset of the CheXpert Plus dataset. From the Table \ref{tab:NLG_Metrics_results} to Table \ref{tab:resource_comparison}, we can find that
Our EMRRG has demonstrated remarkably strong performance in terms of both computational efficiency and predictive accuracy, yielding surprisingly favorable outcomes. It fully validated the efficiency of our proposed framework for the X-ray image based medical report generation.

\subsection{Visualization} 
As shown in Fig.~\ref{fig:visualization}, we present several examples to demonstrate the effectiveness of our proposed EMRRG model for X-ray image-based report generation. For specific X-ray images, we compare the ground-truth reports with those generated by the MambaXray-VL model and our EMRRG model. The selected X-ray images include different views and a variety of conditions, allowing for a more comprehensive and objective evaluation. For a more intuitive visualization, we have highlighted the parts that match the ground truth. The yellow highlighted area represents the part of the report generated by the MambaXray-VL model that matches the ground truth, while the cyan highlighted area corresponds to the part of the report generated by the EMRRG model that matches the ground truth. The pink-highlighted area indicates the portion of the report generated by both models that matches the ground truth. It is evident that the reports produced by our EMRRG model contain a greater amount of content consistent with the ground truth compared to those generated by MambaXray-VL, confirming the superior accuracy and effectiveness of our approach.

\subsection{Limitation Analysis}  
This paper provides an Efficient model for the X-ray image based medical report generation. The LLMs evaluated in this work focus on 7B, which is hardware-friendly, and the LLMs with more parameters are not discussed due to the limited computational resources. On the other hand, due to computational resource constraints, we neither fine-tuned all components within the Mamba architecture nor identified an optimal subset of fine-tuned components.

\section{Conclusion}  
In this paper, we propose an efficient architecture for X-ray image-based medical report generation. This architecture achieves enhanced performance through two methodological advancements: 1) parameter-efficient fine-tuning of the Mamba structure within the MambaXray-VL framework using the $\mathrm{LoRA}_{P}(X)$ method, and 2) partial replacement of conventional decoder layers with Hybrid Decoder Layers in large language models. The effectiveness of our proposed EMRRG model was validated through rigorous comparative evaluations against state-of-the-art medical report generation algorithms on the IU X-ray, MIMIC-CXR, and CheXpert Plus datasets.

\bibliographystyle{splncs04}
\bibliography{reference}

%
%
%
%




\end{document}